\pdfoutput=1

\documentclass[11pt]{article}

\usepackage[]{EACL2023}

\usepackage{times}
\usepackage{latexsym}

\usepackage[T1]{fontenc}

\usepackage[utf8]{inputenc}

\usepackage{microtype}

\usepackage{inconsolata}

\usepackage{times}
\usepackage{latexsym}
\usepackage[T1]{fontenc}
\usepackage[utf8]{inputenc}

\usepackage{microtype}

\usepackage{microtype}
\usepackage{graphicx}
\usepackage{algorithm} 
\usepackage{algpseudocode} 
\usepackage{amsmath} 
\usepackage{amsthm}
\theoremstyle{definition}

\usepackage{bm}

\usepackage{pgfgantt}
\usepackage{natbib}
\usepackage{graphicx}
\usepackage{hyperref}

\usepackage{chngpage}
\usepackage{amssymb}
\usepackage{pifont}
\usepackage{blindtext}

\usepackage{float}

\usepackage[most]{tcolorbox}

\usepackage[utf8]{inputenc}
\usepackage{amsmath}
\usepackage{dsfont}
\usepackage{bm}
\usepackage{soul}

\usepackage{wrapfig}
\usepackage{subcaption}
\usepackage{bbold}

\usepackage{xcolor}		
\usepackage{color, colortbl,soul}
\definecolor{LGray}{gray}{0.9}
\definecolor{Gray}{gray}{0.8}
\definecolor{DGray}{gray}{0.7}
\sethlcolor{DGray}
\definecolor{darkblue}{rgb}{0, 0, 0.5}
\hypersetup{
  colorlinks=true,
  citecolor=darkblue, 
  linkcolor=darkblue, 
  urlcolor=darkblue
}
\newcommand{\squishlist}{
      \begin{list}{$\bullet$}{
           \setlength{\itemsep}{0pt}
           \setlength{\parsep}{3pt}
           \setlength{\topsep}{3pt}
           \setlength{\partopsep}{0pt}
           \setlength{\leftmargin}{1.5em}
           \setlength{\labelwidth}{1em}
           \setlength{\labelsep}{0.5em}}}
\newcommand{\squishend}{
     \end{list}}

\title{Creation and evaluation of timelines for longitudinal user posts}

\author{
\textbf{Anthony Hills$^{1}$, Adam Tsakalidis$^{1,2}$, Federico Nanni$^{2}$, Ioannis Zachos$^{3}$, Maria Liakata$^{1,2,4}$}\\
       $^1$Queen Mary University of London, $^2$The Alan Turing Institute,\\
       $^3$University of Cambridge,
       $^4$University of Warwick\\
      \texttt{\{a.r.hills;a.tsakalidis;m.liakata\}@qmul.ac.uk}
}

\begin{document}
    \maketitle

    
    
    \begin{abstract}
    
    There is increasing interest to work with user generated content in social media, especially textual posts over time. Currently there is no consistent way of segmenting user posts into timelines in a meaningful way that improves the quality and cost of manual annotation. Here we propose a set of methods for segmenting longitudinal user posts into timelines likely to contain interesting moments of change in a user's behaviour, based on their online posting activity.
    We also propose a novel framework for evaluating timelines and show its applicability in the context of two different social media datasets. Finally, we present a discussion of the linguistic content of highly ranked timelines. \footnote{\url{https://github.com/Maria-Liakata-NLP-Group/timeline_selection_and_evaluation}}
    \end{abstract}
    
    
\section{Introduction}\label{sec:introduction}

An increasing body of work considers time-aware models trained on social media data for a number of different tasks, including personal event identification~\cite{li_timeline_2014,li2014major,chang2016timeline}, suicidal ideation and suicide risk detection~\cite{coppersmith2014quantifying,coppersmith2018natural,cao2019latent,matero2019suicide,sawhney2020time,sawhney2021phase}. 
For such tasks deriving meaningful \textit{timelines} (i.e. sequences of posts by individuals), containing examples of the phenomenon under study from large-scale collections, together with associated annotations, is crucial. This is especially important for computational approaches in mental health (MH) given the surging numbers of those seeking help online~\cite{neary2018state}. 

Earlier work on personal life event detection considered selecting salient timelines through topic modelling ~\cite{li_timeline_2014,li2014major} or through a non-parametric generative approach~\cite{chang2016timeline}. However, such approaches are unsuitable for identifying changes in mood or MH more generally. Specifically, since timelines are selected based on linguistic content this introduces a sampling bias for downstream linguistic analysis and annotation~\cite{olteanu2019social, mishra_snap-batnet_2019}. In recent work on suicidal ideation detection, timelines are chosen as the $N$ most recent posts~\cite{sawhney2020time}, which are not necessarily the most salient for annotation. 

\noindent\textbf{Present Work:} We propose a set of methods and associated evaluation framework for identifying salient timelines from the history of social media users to be annotated for changes in a user's behaviour, as revealed through their textual data. Applying our methods in the domain of MH, we follow earlier work in hypothesising that posting behaviour can be a proxy for changes in the MH of an individual~\cite{de_choudhury_discovering_2016}. Therefore we develop methods for creating timelines based on time-series of posting frequency, such as change-point and anomaly detection approaches, and evaluate these against keyword-based methods and randomly selected timelines, in the context of the task of capturing \textit{Moments of Change (MoC)}. A MoC is a  particular point or set of points in time denoting: (1) a shift in an individual's mood from positive-to-negative or vice versa; or (2) a gradual mood progression \cite{tsakalidis2022MoC}. 
We show that our proposed timeline segmentation methods can consistently select timelines that are rich in MoC for large scale cost-effective annotation. We make the following contributions:

\squishlist
        \item We present approaches for extracting timelines from users' posting history on social media based on change-point detection and anomaly detection methods (\S\ref{sec: methods}).
        
        \item We propose a novel evaluation framework for assessing the quality of annotated timelines, and timeline selection methods, which we evaluate on the task of capturing MoCs (\S\ref{Medoid Votes}) on two different social media datasets.
        
        \item We provide a linguistic analysis of timelines obtained, distinguishing timelines dense in MoCs, from timelines sparse in MoCs (see \S\ref{sec: results}). 
        
\squishend
 
    \section{Related Work}\label{sec: related work}

Since we aim to segment users' entire posting history into smaller sequences, manageable to annotate and salient in terms of containing moments of change in mental health, we consider work in the following areas: mental health monitoring (\ref{sec: Tracking Changes in Mental Health}); text segmentation (\ref{sec: text segmentation}); timeline summarization (\ref{sec: timeline summarization}); 
change-point detection (\ref{sec: cpd}). 

\subsection{Tracking Changes in Mental Health (MH)}\label{sec: Tracking Changes in Mental Health}

\noindent{\textbf{Moments of Change (MoC)}} are important in MH tracking.~\citet{10.1145/3290605.3300294} identifies a MoC as a positive change in sentiment for a user with respect to a distressing topic mentioned in a conversation thread. \citet{de_choudhury_discovering_2016} investigated shifts to suicide ideation with models predicting when users transition to posting on a suicide support forum. We consider a more general definition of MoC (\S\ref{sec:introduction}, ``Present Work'').

\noindent\textbf{Creation of Mental Health Datasets.}
A large body of work in creating MH datasets involves labelling posts for symptoms \cite{gkotsis2017characterisation, loveys2017small, cheng2017assessing} or levels of suicide ideation \cite{masuda2013suicide, coppersmith2016exploratory, shing2018expert}.
While annotations for some of these datasets are obtained through proxy signals (e.g., self-disclosure of diagnoses, posts on support networks) questions arise as to how to select appropriate data for annotation. \noindent\citet{mishra_snap-batnet_2019} use keyword based methods to identify posts exhibiting the phenomenon under study (e.g. suicidal ideation) but this leads to sampling biases. 

\subsection{Text Segmentation (TS)}\label{sec: text segmentation}

TS \cite{beeferman1999statistical, pak2018text} focuses on splitting a large body of text (document) into smaller chunks (segments or ``regions of interest'' \cite{oyedotun2016document}). TS has been applied in numerous fields, including emotion \cite{wu2007comprehensive} and sentiment detection \cite{chiru2013sentiment}, often involving segmenting news articles \cite{gao2010sentiment} and review items \cite{sun2013probabilistic}.
While there is some work in segmenting large bodies of social media posts into text segments \cite{kaur2019deep}, we are not aware of work segmenting entire posting histories into smaller, more manageable segments (i.e. timelines),  
to improve downstream longitudinal annotation.

Furthermore, TS primarily operates on linguistic content, rather than timestamped information, with algorithms designed to identify segments containing certain topics of interest, resulting in selection bias \cite{riedl-biemann-2012-topictiling, takanobu2018weakly, hananto2022text}. An alternative is to consider timeline extraction approaches agnostic to the linguistic content, inspired by Timeline Summarisation and Change-Point Detection (CPD).

Evaluation metrics other than precision and recall have been proposed to account for near misses during text segmentation. $P_k$ \cite{beeferman1999statistical} uses a $k$-sized sliding window on a document to compare predicted \textit{vs} ground-truth segmentation locations, assigning partial credit to near misses. 
  However, it is affected by variations in segment sizes and penalizes false negatives more than false positives. WindowDiff  \cite{pevzner2002critique} penalises the latter equally. 
Both metrics require ground-truth annotations of the optimal segmentation locations. 
We propose an approach (\S\ref{sec: evaluation_of_candidate_timelines}) to evaluate segmentation of users' histories based on the proportion of desired annotation labels within a set of sampled sequences of posts (timelines).


\vspace{-.2cm}
\subsection{Timeline Summarization (TLS)}\label{sec: timeline summarization}
TLS aims to provide concise chronologically ordered timelines consisting only of the most relevant information for a given topic or entity, summarizing the key points in time. While TLS has been most commonly applied in news topic summarization \cite{swan2000automatic, martschat_improving_2017, martschat_temporally_2018, steen_abstractive_2019}, there has been increasing interest in applying TLS to social media data~\cite{li_timeline_2014, chen2019learning, ansah2019graph, wang2021bringing}. 

TLS consists of a 2-step pipeline: (1) date selection, then (2) summarisation.
Salient dates summarizing a timeline are typically identified using textual content, as well as time-series information in the history of an individual/topic. Focusing on viral buzzes of celebrity mentions on social media, \newcite{chang_lifecycle_2016, chang2016timeline} aims to select dates by modelling linguistic content and frequency-based time-series patterns. 


\subsection{Change-point Detection (CPD)\label{sec: cpd}}

While CPD has been explored to some extent in news TLS \cite{hu2011generating}, it remains under-explored for social media data.
\noindent{\textbf{Change-points (CPs)}} are defined as points in time where the underlying generative parameters of a data sequence are predicted to have changed \cite{van2020evaluation}. CPD therefore often involves learning a predictive model of a data sequence. In \S \ref{sec: methods}, we use automatically detected CPs to identify salient dates for selecting timelines of users on social media for annotation. While several continuous models exist (e.g. Gaussian ~\cite{adams_bayesian_2007}), we focus on models suited to discrete time-stamped data \cite{pmlr-v80-knoblauch18a} -- such as when posts/comments are made on social media. In such scenarios Temporal Point Processes (TPPs)~\cite{daley2003introduction} are well suited.

\noindent{\textbf{Temporal Point Processes (TPPs)}}
TPPs are stochastic processes that model discrete events localized in continuous time. They are typically characterized by an intensity function, $\lambda$$>$0, which represents the instantaneous rate of event occurrence. 


In order to use TPPs to model event sequences, and predict associated changes -- certain CPD models, such as Bayesian Online Change-point Detection \cite{adams_bayesian_2007} require the TPP to be part of the exponential family of distributions (e.g. Poisson). This is so that the intensity $\lambda$ can be further modelled from a prior conjugate distribution, making it possible to construct the likelihood of the chosen predictive model in a closed form. 
    \section{Approach for Selecting Timelines}\label{sec: methods}
\noindent{\textbf{Task.}} Our principal aim is to select timelines for annotation that are rich in changes in posting behaviour on a MH platform, which we consider as a proxy for changes in MH -- in particular, Moments of Change (MoC). To achieve this, we test a series of timeline selection methods (\S\ref{sec: identify candidate change-point}-\S\ref{sec: timeline creation}), which we evaluate using our proposed framework (\S\ref{sec: evaluation_of_candidate_timelines}). 

\noindent\textbf{Selecting Candidate Timelines}.
To select timelines for annotation, we extract candidate timelines as a span of timestamps $S$ from a user's $u$ history $H$. We first propose identifying changes in posting behaviour as \emph{Candidate Moments of Change} (CMoC), which are dates hypothesised to be surrounded by many MoCs (\S \ref{sec: identify candidate change-point}). Subsequently, we extract the user's posts surrounding these CMoC within a fixed time window, as timelines to be returned for annotation (\S \ref{sec: timeline creation}).

\subsection{Identifying Candidate MoCs (CMoC)}\label{sec: identify candidate change-point}
We investigate the following for identifying CMoC:

\vspace{.1cm}
\noindent\textbf{(1) Bayesian Online Change-point Detection} (BOCPD): 
In a recent evaluation involving experiments with synthetic and real-world change-points,  \citet{van2020evaluation} showed that BOCPD was the best model for a variety of CPD tasks. BOCPD learns a predictive model on a data sequence. When changes in the model's generative parameters are identified, CPs are declared. BOCPD is typically fit with continuous models (e.g. the Gaussian distribution). However, in our case we consider models for discrete event-based data \cite{pmlr-v80-knoblauch18a}.

Since we hypothesize that changes in posting behaviour coincide with changes in mood (see ``Present Work'' in \S1), we use BOCPD to identify changes in individuals' posting frequency. 
As such we consider the daily frequency of posts made by a user as a TPP, and  use the homogeneous Poisson-Gamma (PG) point process model with BOCPD ~\cite{pmlr-v80-knoblauch18a} to fit and identify changes in the daily frequency of posts by a user from their entire associated history. We assess our hypothesis by evaluating timelines obtained this way in terms of how dense they are in MoCs, changes in mood and sentiment (Table \ref{tab:feats_importance}). 

By using a PG model with BOCPD, we assume that each point in a user's posting frequency is sampled from a Poisson distribution with a discrete $\lambda$. Here $\lambda$ represents the expected number of posts by a user within a given time interval. As we use this conjugate Bayesian model, $\lambda$ is  further assumed to be drawn from a Gamma distribution with a set of priors  $\alpha_0$ and $\beta_0$, that act as initial hyper-parameters in our model, where $\alpha_0/\beta_0$, $\alpha_0/\beta_0^2$ denote the prior mean and variance over $\lambda$. BOCPD has an additional hyper-parameter which is the hazard $h_{0}$, where 1/$h_{0}$ expresses a prior belief about the probability of CPs occurring at a given time $t$, provided that a CP has not recently occurred: a low $h_{0}$ results in the over-generation of change-points while a large $h_{0}$ is more conservative and returns very few CPs (ideal in our scenario, to ensure that we do not waste annotation resources, by avoiding annotating too many timelines generated by noise). As such, we experiment with two settings of BOCPD to identify CMoCs: BOCPD (1) and BOCPD (2), which have priors ($\alpha_{0}$:$.01$; $\beta_{0}$:$10$; $h_{0}$:$10^3$) and ($\alpha_{0}$:$1$; $\beta_{0}$:$1$; $h_{0}$:$10$) respectively. 

\begin{figure}
    \centering
    \includegraphics[width=0.5\textwidth]{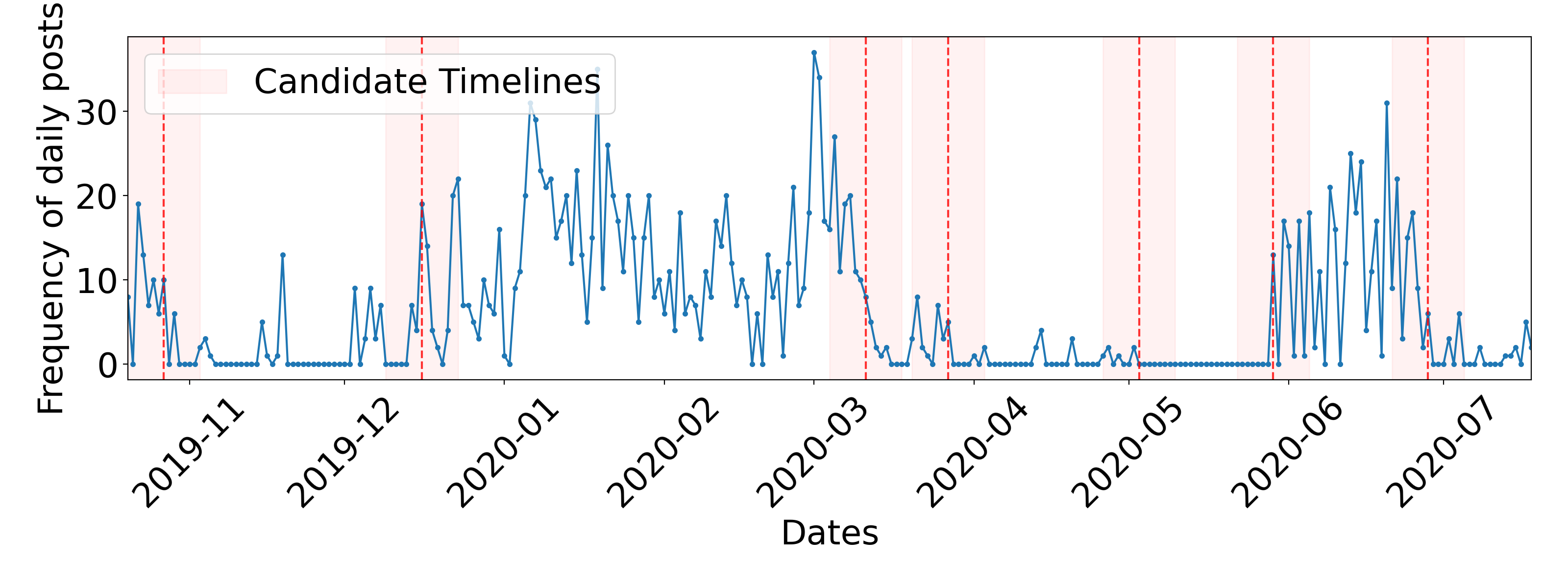}
    \vspace{-.9cm}
    \caption{Using change-points in an example user's posting behaviour to define candidate moments of change $M^{(c)}_{u}$ (dashed red line). Candidate timelines are then created centred on each $M^{(c)}_{u}$, with a radius $r$=7.}
    \label{fig:change-points illustration}
\vspace{-.4cm}
\end{figure}

Since BOCPD computes a full probability distribution over the location of the CPs, quantifying probable CPs along with their associated uncertainty, we use the maximum a posteriori (MAP) segmentation of the probability distribution to return exact point estimates for CPs \cite{fearnhead2007line, van2020evaluation}, which in our setting define CMoCs. An illustration of identifying CMoCs from a given user's history in our implementation of BOCPD is provided in Fig.~\ref{fig:change-points illustration}. 

\vspace{.1cm}
\noindent\textbf{(2) Anomaly Detection (AD)}: Here we aim at identifying (a) days of abnormally high user activity and (b) abnormally long time periods of no user activity at all. We hypothesize that such points in time can be used to select salient timelines. We experiment using different features to fit our model, including the daily frequency of a user's posts and the number of comments they receive for those corresponding posts by others. Using either activity type, we scan over the user's entire history.

For (a) we explore the use of \emph{Kernel Density Estimation (KDE)} \cite{10.1214/aoms/1177728190, scott2015multivariate} to estimate the probability density function of the user's activity. For (b), we focus on time periods in the user's history lasting at least 14 days during which the user had no activity (posts/comments) at all. Given the past 90 days of a user's activity, if the probability on a particular day of seeing either (a) a high volume of activity or (b) a long period of `silence' is lower than $.01$, then we mark the start of this period as an `anomaly' -- i.e., CMoC. We explore (a) and (b) separately for posts and comments, and we also explore concatenating CMoCs identified for high and low posting activity for either comments received or posts made.

\vspace{.1cm}
\noindent\textbf{(3) Keywords}: We incorporate a baseline for identifying CMoCs based on a set of keywords in the \emph{suicide risk severity lexicon} \cite{gaur2019knowledge}. Each keyword present in the lexicon corresponds to different levels of suicide risk severity such as ``I’m tired of this suffering'', and ``I'm going to kill myself''. We hypothesize that the presence of such phrases in a user's post may be indicative of a MoC. This method returns CMoCs for timestamps of posts by a given user that contain a keyword within the lexicon. Note that keyword methods are prone to sampling bias for downstream linguistic analysis, we include them in our experiments due to their popularity for comparison purposes. 


\vspace{.1cm}
\noindent\textbf{(4) Random \& Every day}: We incorporate two na\"ive baselines, as such methods are important for benchmarking in MH tasks~\cite{tsakalidis2018can}. ``\textit{Random single day}'' selects a single date from a uniform distribution over all days in a user's posting history $H$ as a CMoC, $C$ (we evaluate against 100 random seeds to report average scores, \S\ref{sec: evaluation_of_candidate_timelines}). ``\textit{Every day}'' returns every day as a CMoC -- we employ it to see how well our methods are at avoiding the over-generation of candidate timelines. We seek to avoid over-generating timelines as we want to only return timelines with a high density of MoC to improve annotation efficiency.

\vspace{-.2cm}
\subsection{Extracting Posts}\label{sec: timeline creation}

Once a CMoC, $C$, is found, a span of timestamps $S$ from the user's history $H$ is identified within a radius $r$\footnote{Here we take $r=7$ which gives a manageable amount of posts while providing context before and after the CMoC.} around $C$. 
A candidate timeline then consists of the associated sequence of posts, corresponding timestamps and comments within $S$.  
    
\vspace{-.2cm}
\section{Evaluation of Selected Timelines}\label{sec: evaluation_of_candidate_timelines}
While there is previous work in evaluating segments of posts in text segmentation (\S \ref{sec: text segmentation}) and timeline summarization (\S \ref{sec: timeline summarization}), there is little to no prior work on frameworks for evaluating timeline selection methods for the purposes of efficiently annotating longitudinal datasets. As such we identify this as a nascent area of study -- ripe for others to build upon, and propose a novel evaluation framework for selecting timelines for this task. 

We investigate several metrics for evaluating the methods from \S\ref{sec: methods} in terms of their ability to select timelines that correspond to a high proportion of Ground-truth Moments of Change (GTMoC), denoted hitherto as $G$. 
Each CMoC generated by a method as a change point is denoted hitherto as $C$. Since we do not have access to manual ground truth annotations outside of the span of our annotated timelines, we can only evaluate methods according to CMoCs that fall within them.

\subsection{Time-varying Classification Metrics}\label{sec:precision recall f1}

We use the precision and recall metrics by \citet{van2020evaluation} for evaluating change-points (CPs) -- i.e., CPs are evaluated based on the distance $d_{\text{GTMoC}}$ of the predicted CP $C$ falling within a margin of error distance $\tau$ to Ground-truth Moments of Change $G$. 
For our scenario, $\tau$ is reflective of the length of the timelines to be created, and is roughly the radius of a timeline. It should also be chosen based on the uncertainty of the annotation labels. The pros of making assessments based on high performance with a small $\tau$, is that this suggests that very narrow timelines can be created, while still capturing the annotation labels. This allows many timelines to be annotated, thus increasing the diversity of the dataset. However, if timelines are too small, there may not be enough context provided to annotators to perform the annotation task. Thus, allowing for larger timelines provides more context to annotators, which can potentially improve the quality of annotations -- but increase the cost and time to perform the annotation. In our experiments we make assessments based on moderately sized $\tau$ to allow for moderately sized timelines. 
We use $\tau$ = 5 days in table 2, which is the same value used in the experiments of \cite{van2020evaluation}.

A true positive (TP) therefore corresponds to an intersection of a $G$ with a $C$: $G$$\cap$$C$, if $|G-C|$$\leq$$\tau$. We ensure there is a 1:1 mapping between each $G$ and $C$ -- where each $C$ can only intersect  as TP against a single $G$. The total number of TPs for a timeline therefore is given by $\text{max}(|G \cap C|) \leq \text{max}(|G|, |C|)$, where $G$ and $C$ are sets of dates in annotated timelines. The precision and recall are thus defined  as $P = \frac{|G \cap C|}{|C|}$ and $R = \frac{|G \cap C|}{|G|}$, respectively. We compute $P$ and $R$ for each annotated timeline and report mean across all timelines. The mean scores are then used to compute the mean F1.

While these metrics evaluate how well a timeline selection method can identify CMoCs close to GTMoCs, they cannot tell us which method is able to return timelines that contain a high proportion of GTMoCs relative to the number of posts (timelines with high density of GTMoCs). 
Thus we propose an alternative metric (Medoid Votes) based on densities of GTMoCs, as discussed next.

\subsection{Medoid Votes (MV)}\label{Medoid Votes}

We propose a new metric, MV, to account for the inability of prior metrics to consider the density of labels within timelines.
Although a method may have high precision (yielding a prediction close to a ground truth label), the timelines overall may contain a low proportion of the labels that we seek to annotate -- leading to inefficient annotation. Hence, we introduce MV which assigns true positives against \emph{dense regions} of labels as opposed to single labels. As we demonstrate in our experiments, assessments made using MV are more robust, resulting in timelines centered around highly dense regions of the labels we seek to annotate.

To make assessments using MV, first we identify periods in manually pre-annotated user timelines 
that contain a high proportion of GTMoCs relative to the number of posts within the timelines (dense regions) 
(\S\ref{sec: Identifying dense regions in annotated timelines}). We then assign votes to methods that identify CMoCs close to these, and obtain a ranking 
(\S\ref{sec: scoring}). 



\subsubsection{Dense Regions in Annotated Timelines}\label{sec: Identifying dense regions in annotated timelines}


\noindent{\textbf{Medoids.}} We use the notion of `medoids' to represent the location of dense regions of GTMoCs. A \textit{medoid} $M$ is the timestamp of the GTMoC in a given timeline $T$, from which the (Euclidean) distances $d(. , .)$ of all other timestamps of annotated GTMoCs $G$ in timeline $T$ are minimal:
\begin{equation}
 \label{eq:medoid}
   M = \underset{G_{a} \in T}{\arg\min} \sum_{G_{b} \in T} d(G_{a}, G_{b})  
 \end{equation}
 
 \vspace{-.2cm}

\noindent{\textbf{Density of annotated timelines.}} 
We further characterise the locations of dense regions (medoids) by the number of GTMoC they contain. This \textit{`density'} of a timeline is defined as
$  \rho = \frac{|G|}{|p|}$, where $|G|$ is the sum total number of GTMoCs within an annotated timeline $T$ and $|p|$ is the number of posts in $T$. 
 
 In order to weight timelines by how dense they are in GTMoCs, a medoid $M$ inherits the density $\rho$ of the timeline $T$ it represents. We transform $\rho_{T}$ for each $T$, to provide a binary distinction between ``dense'' (+1) and ``sparse'' (-1) medoids as:
\begin{equation}\nonumber
\label{eq:density}
  \rho_{T}^{(\text{binary})} 
\begin{cases}
    +1 & \text{if } \rho_{T} \geq \text{Median}(\rho_{T} \ \forall \  T)\\ 
    -1 & \text{otherwise}
\end{cases}
\end{equation}

 A good timeline is therefore one that is ``dense'', and the ideal location for a CMoC is as close as possible to a dense medoid $M$ (see eq.~\ref{eq:medoid}).

In an ideal scenario where we have the resources to annotate many timelines sampled from many candidate methods, we could compare and rank the methods based on the number of dense timelines or the average resulting densities. Alternatively, we could evaluate the proposed methods against a set of fully-annotated user histories. However, due to the high cost and time-consuming process of annotation, such approaches are infeasible. Instead we propose an alternative solution that does not require annotating all the timelines that would be generated (or entire user histories). We do this via a scoring system based on distances of CMoC relative to dense medoids in a small set of trial annotated timelines, as described next.

\subsubsection{Scoring Timeline Selection Methods}\label{sec: scoring}

We employ the evaluation framework in \S\ref{sec: Identifying dense regions in annotated timelines} to assess pre-annotated timelines against CMoCs in timelines selected by different methods. Assuming an annotated timeline $T$, we aim to assess how close an identified CMoC $C$ is to a dense region of GTMoCs within $T$. We therefore give preference to methods that identify CMoCs in close proximity to medoids that are dense in GTMoC, while also penalizing methods that over-generate CMoC. 
\vspace{.1cm}
\noindent\textbf{Distance Scores}.
We calculate the proximity of CMoCs predicted by a method to $M$ as the minimum absolute distance $d_{m}$ (in days) between all CMoCs predicted by a given method $m$ (\S\ref{sec: identify candidate change-point}) for a user's entire history. Then, we compute a distance score for each $m$ per annotated timeline as: 
\begin{equation*}
D_{m} = (d_{m} + \epsilon) * \text{sign}( \rho_{T}^{(\text{binary})}),
\end{equation*}

\noindent where $\epsilon$=.001, to preserve the sign of each medoid's $\rho_{T}^{(\text{binary})}$ in the case of $d_{m}$= 0. $D_{m}$ is then used to denote the proximity of CMoCs predicted by method $m$ (in days) to a ground truth medoid $M$ with density $\rho_{T}^{(\text{binary})}$. Since we want to obtain timelines that are close to dense regions in GTMoC, we seek to identify methods with low positive $D_{m}$.

\vspace{.1cm}
\noindent\textbf{Votes.} To reward methods that identify a CMoC in close proximity to a `dense' $M$ (low positive $D_{m}$), and penalize methods which over-generate CMoC (e.g., in locations that contain a low density of GTMoC), we assign votes to each method $m$ by:
\begin{equation*}
v_{m} = 
        \begin{cases}
            +1 & \text{if } 0 \leq D_{m}  \leq \tau\\ 
            0 & \text{otherwise}\\
        \end{cases}
\end{equation*}
\noindent where $\tau$ is the same margin of error (in days) described in \S\ref{sec:precision recall f1}. This gives a positive vote to a method generating a CMoC that falls within a margin of $\tau$ days to a dense medoid. Votes $v$ are then normalized per timeline and method ($V_{m} = \frac{v_{m}}{|C|}$, where $|C|$ is the total number of CMoCs generated by $m$, that fall within each annotated timeline).

\begin{figure*}
\centering
  \includegraphics[width=.85\textwidth]{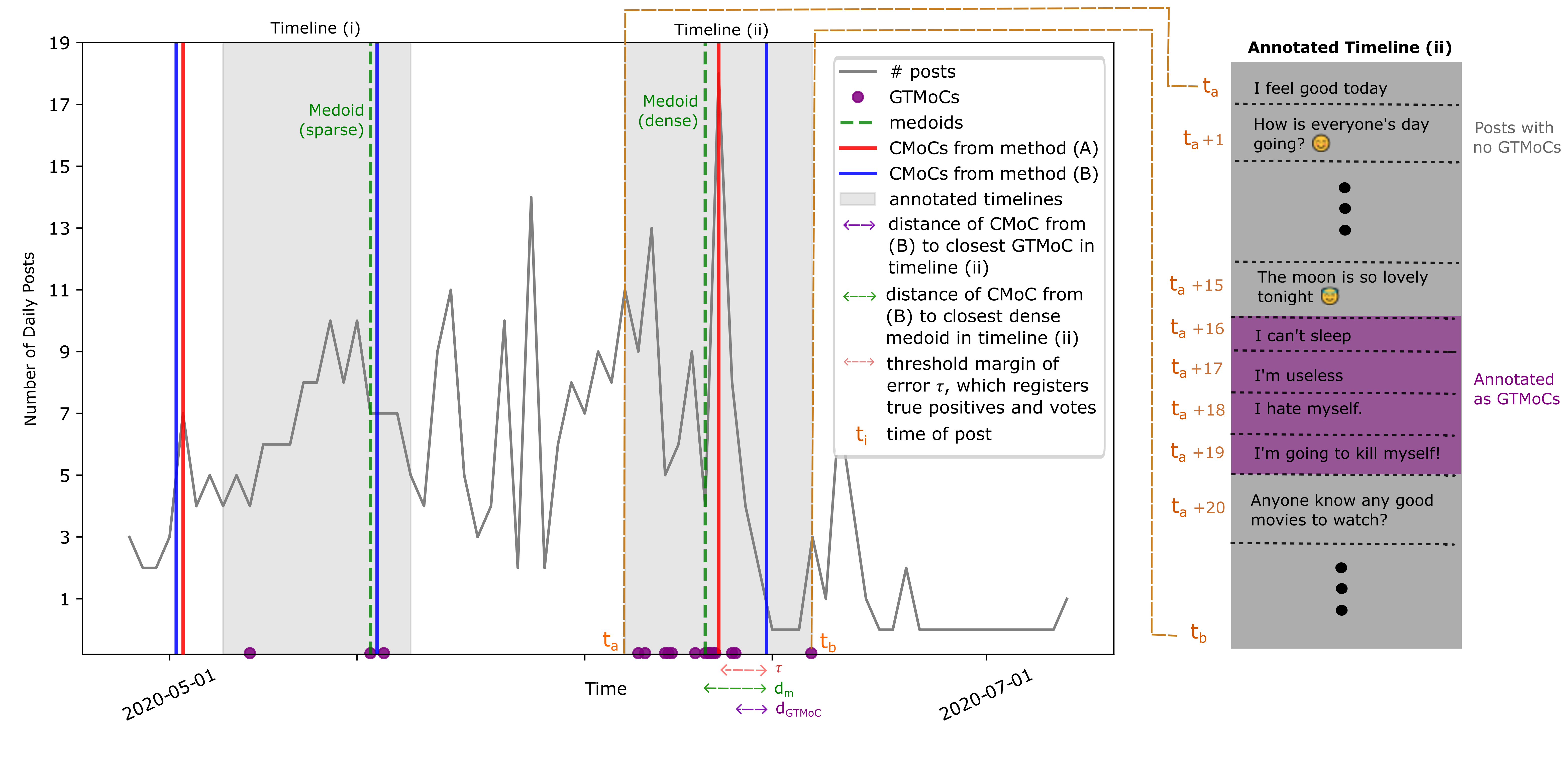}
  \vspace{-.7cm}
    \caption{Evaluation of CMoCs against GTMoCs. Votes and true positives are assigned based on distances $d$ of CMoCs falling within a margin of error $\tau$ against dense medoids or GTMoCs. Here, method A (red) selects better timelines than method B (blue), as these are close to dense regions of GTMoCs ($d_{\text{m}}$$\leq$$\tau$) and labels ($d_{\text{GTMoC}}$$\leq$$\tau$).}
    \label{fig:evaluation extra figure}
    \vspace{-.4cm}
\end{figure*}

\vspace{.12cm}
\noindent\textbf{Scoring of methods}.
Timeline selection methods are subsequently scored and ranked by summing the votes $V_{m}$ for each method $m$ over all $T$. As we are concerned with ranking methods, we then min-max scale our results in the range of 0 to +1, where methods that have scores close to 1 rank near the top and methods that score close to 0 are the worst in their ability to return timelines containing a high proportion of GTMoCs. The scoring of methods proposed in \S \ref{sec: timeline creation} are shown in Table \ref{tab: final results}, and Fig. \ref{fig:final_results} for varying margin of error, $\tau$.
The evaluation framework is visualised in Fig.~\ref{fig:evaluation extra figure}.

    \section{Experiments}

We evaluate our timeline selection methods (\S \ref{sec: methods}), using our evaluation framework (\S \ref{sec: evaluation_of_candidate_timelines}) based on ground-truth human annotated data.

\subsection{Datasets}\label{sec: final dataset} 

We evaluate our automatic timeline selection methods using two datasets (summarised in Table~\ref{tab: datasets summary stats}) from different platforms: The \textit{TalkLife} dataset contains timelines automatically selected using one of our proposed methods. While our evaluation is designed to allow alternative methods to achieve higher scores than the methods used to select timelines we still want to exclude any possibility of inherent bias. To this effect we also evaluate against timelines manually selected from \textit{Reddit} independently from this work ~\cite{tsakalidis2022overview}.

\noindent{\textbf{TalkLife}}\footnote{\url{https://www.talklife.com}} is a peer-support social network  operating primarily as a mobile app. Users are mainly English speakers, 70\% of whom are 15-24 years old \cite{sharma_engagement_2020}. The posts/comments on TalkLife focus primarily on MH, daily-life issues and feelings. It is thus suited to identifying MoC and computationally analysing MH \cite{10.1145/3290605.3300294, sharma_computational_2020, saha2020causal, kim2021you}. We select timelines on the basis of timestamped user posting frequency, and associated comments received.
The context of posts is only used in annotating the selected timelines; thus, methods for timeline selection are transferable to other platforms. 

We licensed a de-identified dataset from TalkLife consisting of 1.1M users (12.3M posts, Aug'11-Aug'20).
Due to the high variance in users' posting frequency, only timelines having [10-150] posts were considered for annotation. This was so that timelines were not impractically long while still providing enough context for annotators to observe and mark a change. 
The final annotated dataset consists of 500 timelines (see Table~\ref{tab: datasets summary stats}), with a mean of 35 posts ($\pm22$). These timelines were selected using BOCPD PG (1), where the parameters ($\alpha_{0}$:$.01$; $\beta_{0}$:$10$; $h_{0}$:$10^3$) were fixed on the basis of improved model performance on a validation dataset of 70 manually annotated timelines selected via anomaly detection. 
All 500 timelines within the evaluation dataset were manually inspected and filtered according to the details in \ref{appendix: ground_truth_timelines}.


\noindent\textbf{Reddit}. 
We further tested the generalizability of our methods and evaluation framework on a different dataset, that was not generated using automatic timeline selection approaches -- the CLPsych 2022 Shared Task corpus \cite{tsakalidis2022overview}. We chose to include this additional dataset to address potential concerns that experiments and analysis performed on the TalkLife timelines have some bias towards the BOCPD method in experiments evaluated on the TalkLife timelines -- as they were selected using BOCPD. 
This corpus was sourced from Reddit, a social media platform where individuals make public posts and which has been studied extensively as a resource for mining textual data for MH studies \cite{de2014mental, losada2016test, shing2018expert, zirikly2019clpsych,  losada2020overview, low2020natural}. We make use of the `Reddit-New' dataset of the CLPsych 2022 corpus, consisting of 139 timelines where 17-82\% of posts come from MH subreddits and had been pre-selected manually by two researchers independently as likely to contain a high proportion of MoCs.  


\vspace{.12cm}
\noindent\textbf{Annotation of GTMoC}\label{annotation guidelines} 
in TalkLife timelines was performed by 3 English speaking (1 native), university educated annotators. Reddit timelines were annotated by 4 English (2 native) speakers 
\cite{tsakalidis2022overview}. 

Annotators were provided with timelines containing chronological posts by users with their associated comments and timestamps. They were asked to label posts containing a `Switch' (sudden change in mood) or an `Escalation' (gradual mood progression) -- a (default) label of `None' was assigned to posts with no MoC. A `Switch' is defined in the guidelines as `a drastic change in mood, in comparison with the recent past', with annotators having to label its beginning and its range.
An `Escalation' is  `a gradual change in mood, which should last for a few posts'. Annotators  had to label the peak of an escalation and the range of associated posts (see Fig.~\ref{fig:switch escalation} of \ref{appendix: annotation guidelines} as an example).

To obtain GTMoC for our evaluation we aggregate the annotations across all annotators per timeline in the same way as \cite{tsakalidis2022MoC}. 
Due to the challenging and subjective nature of the annotation task, the percent of inter-annotator agreement for the labels `None', `Switch' and `Escalation' were .89, .30, and .50 respectively for the TalkLife dataset, and .83, .26, and .31 respectively for the 2022 CLPsych Corpus, based on majority agreement. We consider all labels of `Switch', `Escalation', and their corresponding ranges as GTMoC. We thus merge both labels to define GTMoCs, as we are interested in identifying timelines that contain both types of changes in mood. 

\begin{table}[!h]
\centering
\resizebox{.9\linewidth}{!}{
    \begin{tabular}{|r|r|r|r|c|}
    \hline
        & Timelines & Posts & Users & Timeline Length \\ \hline
        TalkLife & 500 &18,702 & 500 & $\leq$ 2 weeks 
        \\
        Reddit & 139 & 3,089 & 83 & $\sim$ 2 months
        \\ \hline
    \end{tabular}
    }
    \vspace{-.1cm}
    \caption{Summary of datasets used in our experiments.} 
    \label{tab: datasets summary stats}
    \vspace{-.4cm}
\end{table}


\subsection{Results \& Discussion}\label{sec: results}
We identify CMoCs (\S \ref{sec: identify candidate change-point}) on annotated timelines from TalkLife and Reddit (\S \ref{sec: final dataset}), and evaluate using our metrics (\S \ref{sec: evaluation_of_candidate_timelines}). 
We round CMoCs to the nearest day,
de-duplicating dates,
to compare methods.

\begin{figure}
    \centering
    \includegraphics[width=.75\linewidth]{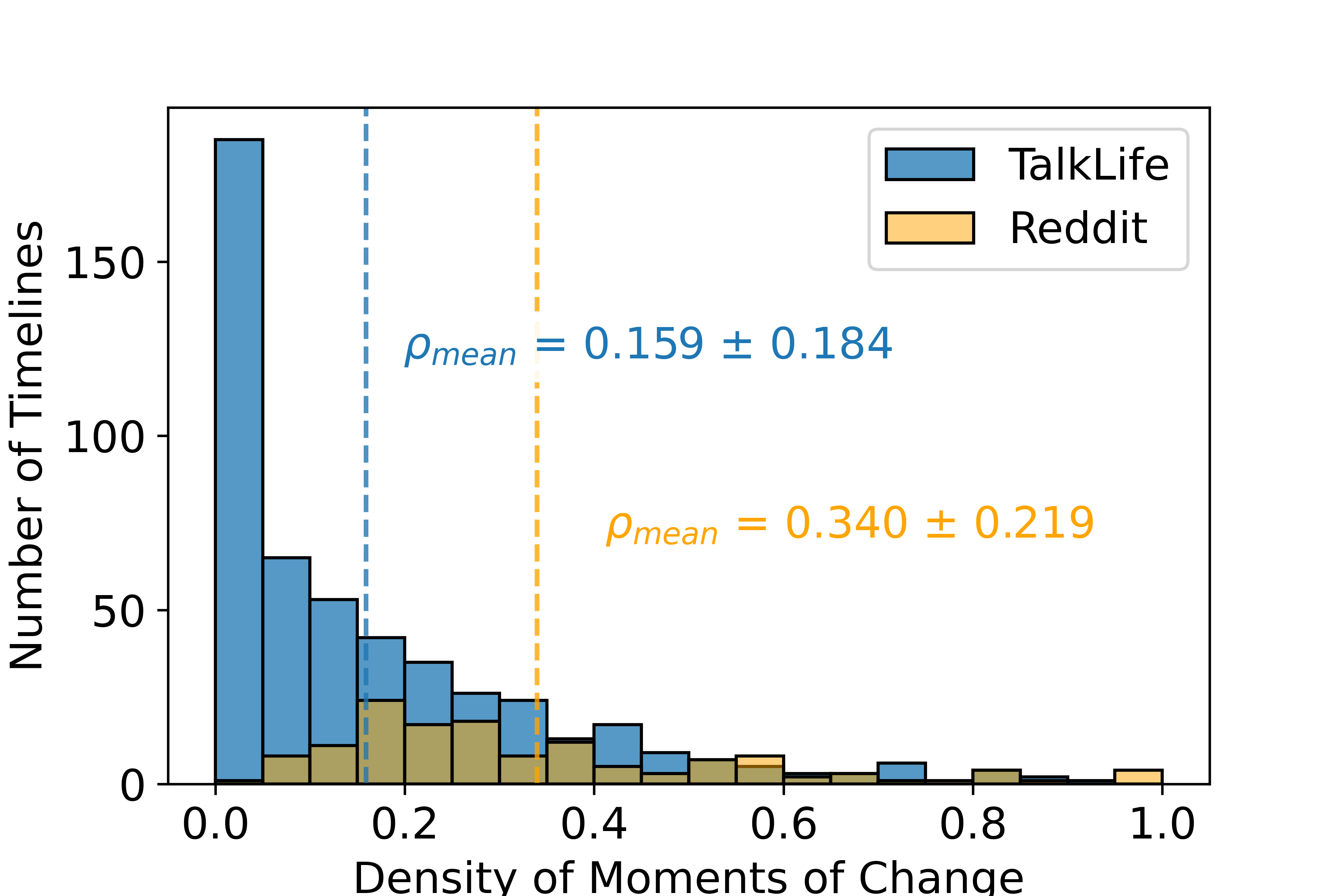}
    \vspace{-.1cm}
    \caption{Density of GTMoCs per timeline. 
    }
    \label{fig:histogram_density_of_moc}
    \vspace{-.4cm}
\end{figure}

\noindent{\textbf{Density scores of annotated timelines.}} The density of the annotated timelines from TalkLife are presented in Fig.~\ref{fig:histogram_density_of_moc}. The mean density (.159) is comparatively high considering that GTMoCs are rare events, and many timelines do not contain any GTMoC. 
While the mean density (.340) of manually selected timelines from Reddit is higher, extra annotation effort was taken by annotators to ensure these timelines had a high proportion of GTMoCs. 

\begin{figure*}
    \centering
    \includegraphics[width=.85\textwidth]{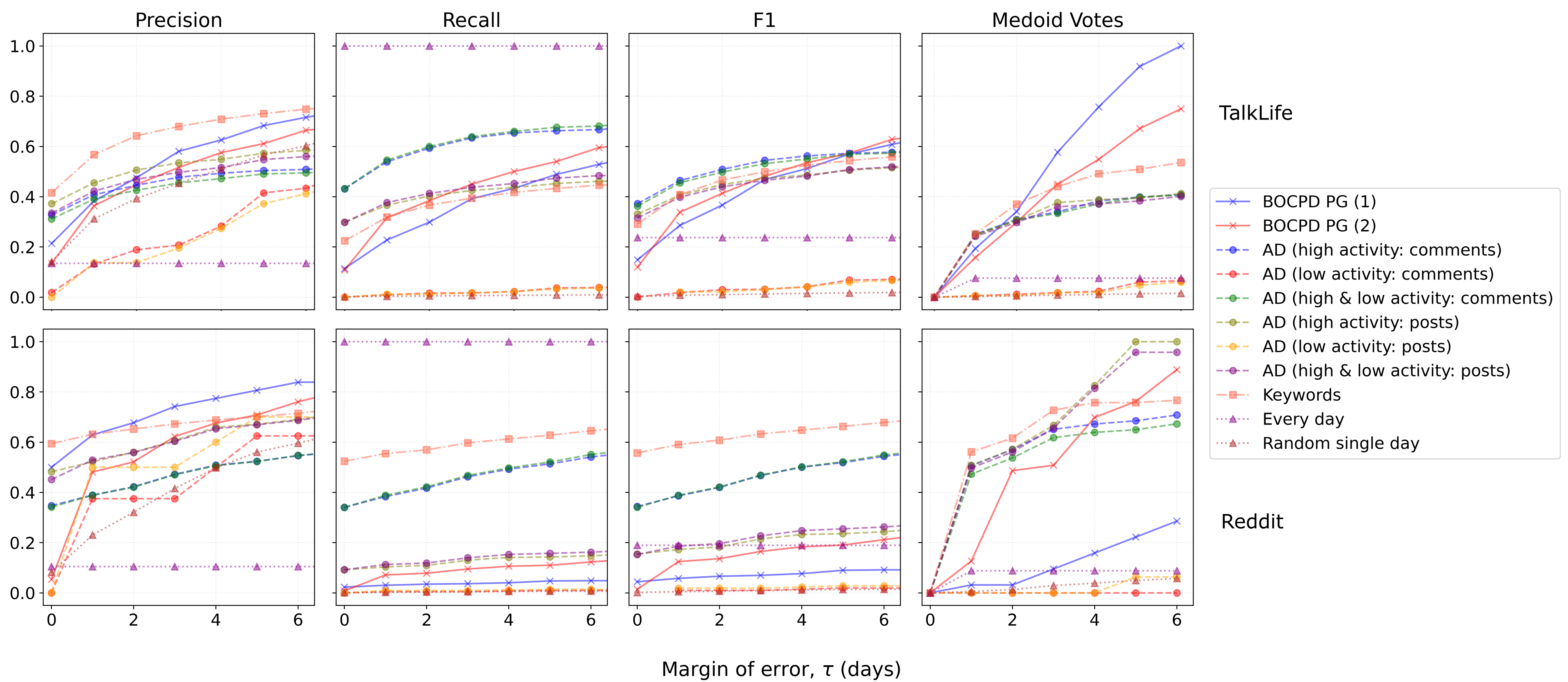}
    \vspace{-.3cm}
    \caption{Evaluation metrics for different timeline selection methods, with varying margins of error $\tau$ (days). 
    }
    \label{fig:final_results}
\end{figure*}

\noindent{\textbf{Ranking of timeline selection methods.}}  Table~\ref{tab: final results} and Fig.~\ref{fig:final_results} shows the generalizability of our models and evaluation based on the consistency of results across both datasets. Overall, BOCPD models achieve the highest precision, and relatively high medoid votes (MV) across varying values of $\tau$. Note that BOCPD PG (1) had hyper-parameters that were tuned for the data on TalkLife, whereas BOCPD PG (2) has very general hyper-parameters – not tuned for either TalkLife or Reddit. Despite not having any models tuned specifically for Reddit, BOCPD (1) achieves the highest precision for the majority of margins of error $\tau$, and BOCPD (2) achieves the 2nd highest precision for larger
$\tau$. Importantly, BOCPD achieves the highest precision for most
cases of $\tau$ across both datasets. Precision is particularly important as it ensures that the resulting CMoCs will have a high chance to be close to
GTMoCs. This aligns with our objective of ensuring the resulting dataset will
be annotated with a high proportion of GTMoCs.

For both Reddit, and TalkLife, the more general parameters of BOCPD PG (2), which were not tuned for either dataset, still achieve among the highest precision and MV 
(next highest MV -- and also the highest $P$ for TalkLife). Even with low $h_0$ and $\alpha_0$/$\beta_0=1$ (likelier to over-generate CMoCs), BOCPD (2) outperforms all AD and na\"ive methods on MV and F1 on TalkLife.  For TalkLife, AD (high activity: posts) achieves slightly worse MV compared to keywords, but outperforms it on Reddit, despite being potentially disadvantaged by not using linguistic content.
AD (low activity) achieve among the worst F1 and MV. As a result, timelines created around anomalously low post frequency would be unsuitable for selecting dense timelines.

Scores vary with $\tau$ (Fig.~\ref{fig:final_results}). For low margins ($\tau$<3) BOCPD ranks lower in F1 and MV in both datasets, but ranks among the highest for larger $\tau$. We attribute this to BOCPD assigning CMoCs to 
transitions
from high to low posting activity.  As we expand $\tau$ and select longer timelines around CMoCs, BOCPD is able to capture moments in time which can contain both high and low posting activity. Transitions from high to low posting activity may not be captured for low $\tau$ -- potentially explaining why the performance in this case is lower than methods that favour a high amount of posts. Since timelines on TalkLife were created with a radius of 7 in~\cite{tsakalidis2022MoC}, setting a fairly large $\tau$=5 is suitable for assessing which methods are able to select dense timelines, while also allowing us to identify  shorter, denser, timelines from longer annotated timelines, as in the case of Reddit.

While recall and F1 are relatively low for BOCPD across both datasets, we argue that precision and MV are the most important metrics to focus on for our task. Considering that `everyday' has a perfect recall of $1.00$, and that annotating all posts in a users history would indeed return all the GTMoCs for a user -- this is highly inefficient and infeasible, and goes against our original objective of \emph{efficiently} annotating a user's posts. By instead focusing on methods with high precision and MV, rather than recall, we ensure that the resulting timelines are near a high proportion of the labels we aim to annotate. This allows annotators to consider fewer posts to capture the same amount of rare labels, which are costly to annotate.

\begin{table}[!h]
\centering
\resizebox{\linewidth}{!}{
    \begin{tabular}{|l|c|c|c|c||c|c|c|c|} 
        \cline{2-9}
    \multicolumn{1}{c|}{} &
      \multicolumn{4}{c||}{TalkLife} &
      \multicolumn{4}{c|}{Reddit} \\
        \hline 
        Method & $P$ & $R$ & $F1$ & $MV$
        & $P$ & $R$ & $F1$ & $MV$
        \\ 

        \hline \hline
        
        BOCPD PG (1)            & \colorbox{LGray}{\textbf{.683}}        &  .489     & \textbf{.570}      &  \colorbox{DGray}{\textbf{.919}}  & \colorbox{DGray}{\textbf{.806}} & .048          & .090          & .222  \\ 
        BOCPD PG (2)            &  \textbf{.611}       & .540      & \colorbox{DGray}{\textbf{.574}}      &  \colorbox{LGray}{\textbf{.672}}  & \colorbox{LGray}{\textbf{.708}}          & .110          & .190          & \textbf{.762}  \\ 
        AD (high comments)      & .504        & \textbf{.662}      & \colorbox{LGray}{\textbf{.573}}      &  .399           & .524          & .513          & \textbf{.519}          & .685  \\ 
        AD (low comments)       & .415        & .037      & .068      &  .060           & .625          & .010          & .020          & .000   \\ 
        AD (high \& low comments)& .491        & \colorbox{LGray}{\textbf{.677}}      & .569      & .399           & .523          & \textbf{.521}          & \colorbox{LGray}{\textbf{.522}}          & .650  \\ 
        AD (high posts)         & .573        & .453      & .506      &  .395           & .671          & .143          & .236          & \colorbox{DGray}{\textbf{1.00}}  \\ 
        AD (low posts)          & .372        & .033      & .060      & .048            & \textbf{.700}          & .014          & .028          & .064  \\ 
        AD (high \& low posts)   & .548        & .474      & .508      & .383           & .669          & .157          & .255          & \colorbox{LGray}{\textbf{.958}} \\
        Keywords                & \colorbox{DGray}{\textbf{.731}}        & .433      & .544      & \textbf{.509}   & .702          & \colorbox{LGray}{\textbf{.628}}          & \colorbox{DGray}{\textbf{.663}} & .758   \\ 
        Every day               & .135        & \colorbox{DGray}{\textbf{1.00}}      &  .237     & .076  & .105          & \colorbox{DGray}{\textbf{1.00}} & .190           & .088  \\
        Random single day       & .567        & .009    & .017      & .014              & .560          & .007          & .014          & .050  \\ \hline
    \end{tabular}
    }
    \vspace{-.3cm}
    \caption{\label{tab: final results} Evaluation of timeline selection methods,
     using a margin of $\tau$=5 days. MV (\S \ref{Medoid Votes}) are min-max scaled in the range $\tau$=[0,6] days. 
     \colorbox{DGray}{\textbf{First}}, \colorbox{LGray}{\textbf{{second}}}, and \textbf{third} highest scores are highlighted.}
     \vspace{-.4cm}
\end{table}

\noindent\textbf{Linguistic analysis of timelines.}
To gain insights into the characteristics of `dense' vs `sparse' timelines, we employ VADER \cite{hutto2014vader}, assigning a sentiment score per post, and Twitter-RoBERTa-emotion \cite{barbieri2020tweeteval}, assigning four emotion scores (joy, anger, sadness, optimism) per post on the TalkLife dataset.
We equally split 250 TalkLife timelines, between `dense' (density $\rho_{u,i}$ is in upper-quartile of all timelines) and `sparse' (bottom-quartile).
The distribution of sentiment scores across these timelines are shown in Fig.~\ref{fig:sentiment_distribution}.
For each timeline we extract statistical features (avg, std, min, max) for each emotion/sentiment dimension of its posts, and the same features based on their difference across two consecutive posts in the timeline.
Using these features, we train a Logistic Regression aiming at predicting `dense' vs `sparse' timelines and extract the coefficients with the highest/lowest values. 

Sparse timelines frequently consist of positive posts in sentiment/mood (see Table~\ref{tab:feats_importance}). On the other hand, sadness- and variance-based features correlate the most with predicting a timeline containing many MoCs -- a finding that was empirically confirmed via manual inspection of the most dense timelines. Developing methods that account for the variability in a user's mood/sentiment is a potential future direction in this regard. 


\begin{minipage}{.96\linewidth}
  \begin{minipage}[b]{0.58\linewidth}
    \centering
    \includegraphics[width=.85\linewidth]{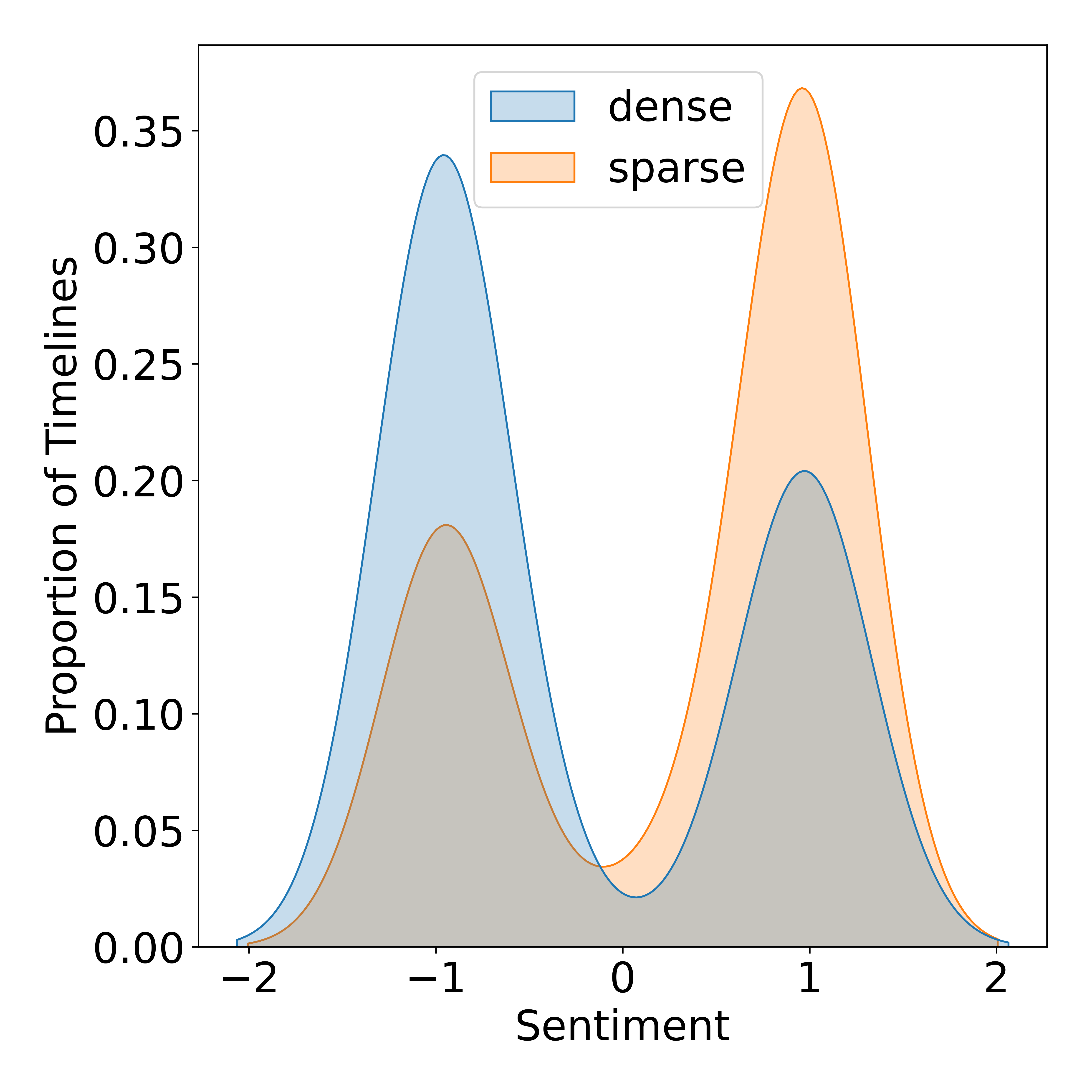} 
    \vspace{-.5cm}
    \captionof{figure}{Sentiments of `dense' \textit{vs} `sparse' timelines (medians: $-.949$ \& $.970$, respectively).}
    \label{fig:sentiment_distribution}
  \end{minipage}
  \hfill
  \begin{minipage}[b]{0.38\linewidth}
    \centering
    \resizebox{.88\linewidth}{!}{%
    \begin{tabular}{ |l|r| } 
        \hline
        Feature & Coef \\
        \hline \hline
        sadness (avg) & 2.29\\
        sadness (std) & 1.45\\
        sentiment (std) & 1.00\\ \hline
        sentiment (avg) & -1.23\\
        optimism (avg) & -1.25\\
        sentiment (min) & -1.31 \\
        joy (avg) & -1.58 \\ \hline
        \end{tabular}%
}
        \vspace{-.2cm}
      \captionof{table}{Logistic Regression coefficients classifying timelines as `dense' (1) or `sparse' (-1).}\label{tab:feats_importance}
    \end{minipage}
  \end{minipage}

\section{Conclusions \& Future work}\label{sec: conclusions}

We have introduced methods and an evaluation framework for identifying timelines from users' social media posts, likely to contain a large amount of Moments of Change (MoC).
We use changes in posting behaviour as a proxy for changes in mood, to 
efficiently identify 
longitudinal user content worth annotating. Our methods have been manually evaluated against ground truth MoCs (GTMoCs) in two different datasets. Bayesian Online Change Point Dection (BOCPD) shows promise in detecting timelines rich in GTMoCs.

Future work can explore the incorporation of textual content in the BOCPD Poisson-Gamma model for the distinction between different types of GTMoC. We find that resulting timelines dense in GTMoCs are characterised by a high deviation in sentiment from one post to the next, suggesting that such deviations may be a useful feature for distinguishing between different types of GTMoC. 

We expect that the methods proposed in our work will benefit researchers interested in creating longitudinally annotated textual datasets of user posts, particularly when annotating Moments of Change. 
\newpage
\section*{Ethics Statement}\label{sec: ethics statement}

Ethics IRB approval was obtained from the Biomedical and Scientific Research Ethics Committee of the University of Warwick (ref: BSREC 40/19-20) prior to engaging in this research study. 
Our work involves ethical considerations around the analysis of user generated content shared on a peer support network (TalkLife). A license was obtained to work with the user data from TalkLife and a project proposal was submitted to them in order to embark on the project. 
The current paper focuses on the identification of periods of interest within the user history, in terms of moments of change. The work on annotation of moments of change (MoC) is separate to this paper but considers sudden shifts in mood (switches or escalations).
Annotators were given contracts and paid fairly in line with University pay-scales. They were alerted about potentially encountering disturbing content and advised to take breaks during annotation. The annotations are used to evaluate the work of the current paper, which aims to meaningfully segment timelines in terms of containing likely moments of change.
Potential risks from the application of our work in being able to identify moments of change in individuals' timelines are akin to the identification of those in earlier work on personal event identification from social media and the detection of suicidal ideation. Potential mitigation strategies include restricting access to the code base and annotation labels used for evaluation.
No data can be shared without permission from the platform or significantly paraphrased. Any examples used from the users' history are anonymised and paraphrased.

\section*{Limitations}
In this work we focus on returning timelines rich in Ground-truth Moments of Change (GTMoCs) in mood, using 
posts on social media which are by definition sparse. This has several limitations. Firstly, our labels of GTMoCs rely on individuals self-disclosing related information. 
We cannot make assessments based on someone's experience offline. The users chosen in our sample may also be users who are more likely to disclose information and so their posting patterns may not be typical of the general population. Both of these issues are true for most work in affective computing from social media.

Our methods for identifying Candidate Moments of Change (CMoCs) have several limitations. Similar to the issues with our GTMoCs, these methods rely on posting behaviour and cannot capture behaviour outside the user's social media history. 
Another limitation of our methods for identifying CMoCs is that they currently only use simple univariate features (e.g. posting frequency), and do not model the influence of cross-user interactions or multivariate features. While we suspect these methods for identifying CMoCs could be extended to model these more complex types of features and interactions, to better select timelines, we have not done this in the current work.

Finally, while we have shown that our methods for identifying CMoCs to select timelines rich in GTMoCs in mood generalize well between two social media platforms (TalkLife and Reddit), we have not experimented with other platforms.
Our methods have been used for returning timelines rich in ground-truth labels for changes in mood but it remains to be seen whether they generalize well to identifying timelines rich in other labels for other related annotation tasks (e.g. labelling levels of suicide ideation). We believe this to be the case. 

\section*{Acknowledgements}
This work was supported by a UKRI/EPSRC Turing AI Fellowship to Maria Liakata (grant EP/V030302/1) and the Alan Turing Institute (grant EP/N510129/1). The authors would like to thank Julia Ive, Theo Damoulas, the anonymous reviewers and the meta-reviewers for their valuable feedback.  

\bibliography{eacl2023}
\bibliographystyle{acl_natbib}

\appendix
\appendix
\section{Appendix}\label{sec:appendix}

\subsection{Creating Ground-truth Timelines, by Retaining a Subset of Representative Candidate Timelines}\label{appendix: ground_truth_timelines}

\begin{figure}
    \centering
    \includegraphics[width=0.5\textwidth]{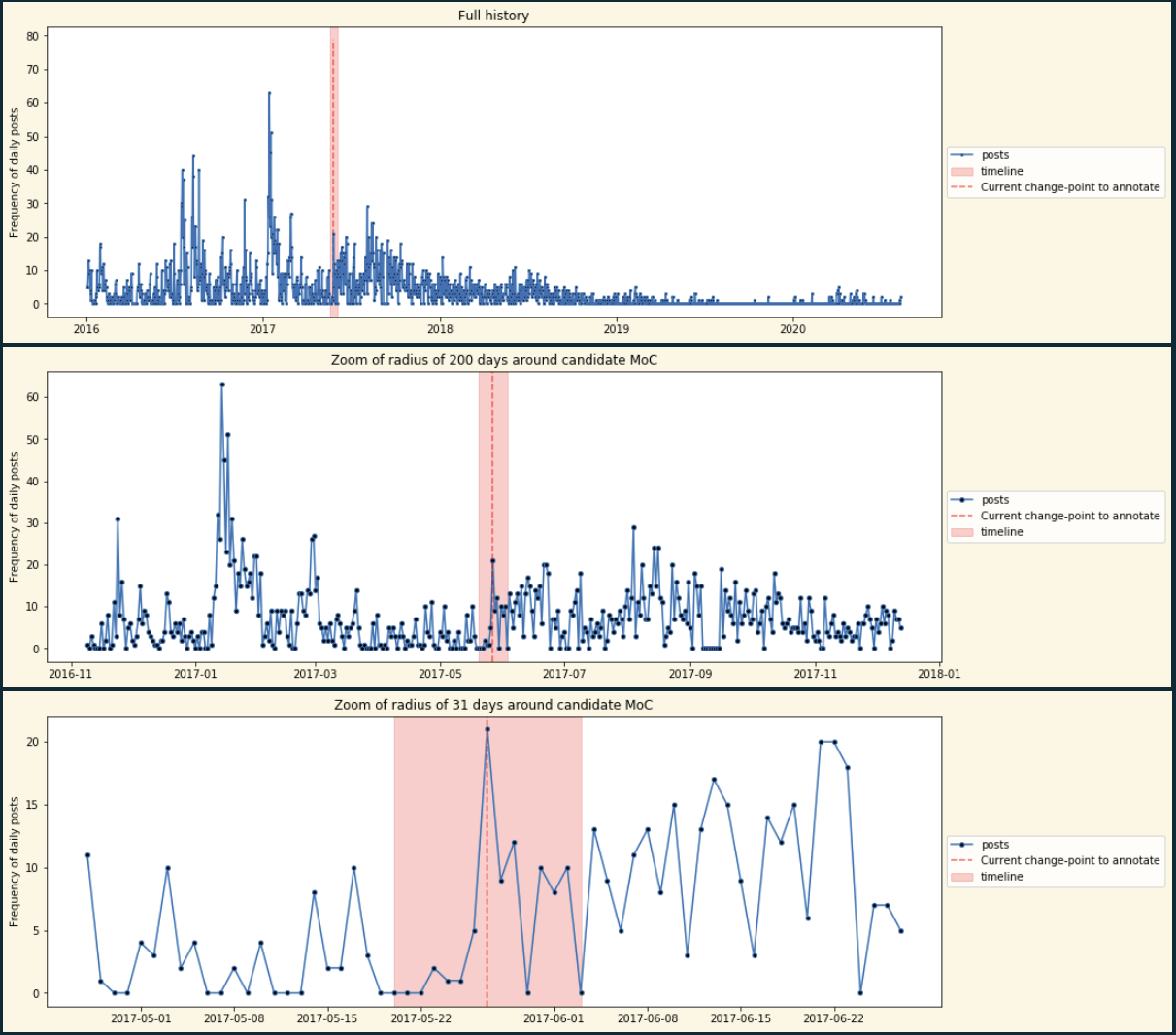}
    \caption{A timeline that was retained, out of the 1,220 timelines manually observed. It was retained as it (1) was not primarily sparse as it contains posts distributed well over the timeline, and (2) appeared to be generated by a plausible change-point rather than noise. Timelines were visualized on 3 time-scales, as shown in this figure, to allow for closer inspection and to compare in context of the full time-series.}
    \label{fig:retained_timeline}
\end{figure}
\begin{figure}[!h]
    \centering
    \includegraphics[width=0.45\textwidth]{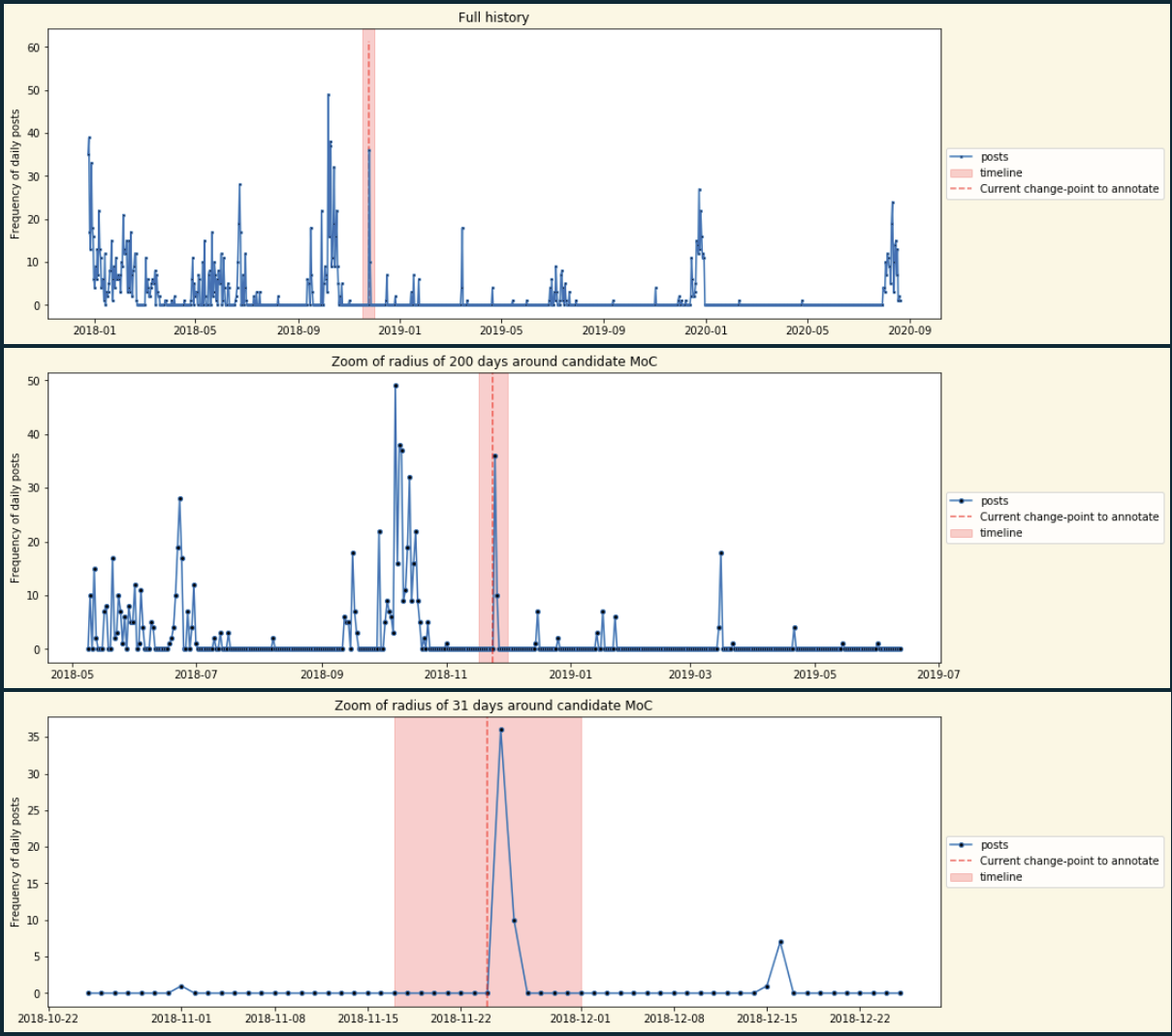}
    \caption{A timeline that was discarded, out of the 1,220 timelines manually observed. It was discarded as it (1) was primarily sparse containing only posts on a few days in the timeline, and (2) appeared to be generated by noise rather than by a realistic change-point.}
    \label{fig:discarded_timeline}
\end{figure}

In addition to the details provided in section \ref{sec: methods}, for selecting candidate timelines, we provide some additional details inline below. As multiple timelines will typically be returned for each user using methods in \ref{sec: methods} and annotating all of these can be time-consuming, in order to keep the 500 annotated ground-truth timelines relatively diverse in terms of the types of users -- only a single timeline was returned per user to be annotated. Therefore, for each user only a single timeline was randomly sampled per and these were presented visually in turn to the first author of this paper, with multiple time-scales limiting the x-axis of the visualization returned: (1) the time-scale of the whole user's history, (2) a radius of 200 days surrounding the CMoC and (3) a radius of 31 days around the CMoC. This was to ensure that the candidate timelines could be inspected in close detail (3), and also observing the timeline in context of the full time-series (1) for that user. These three multiple time-scales for a single user are presented visually in figure \ref{fig:retained_timeline}. A manual binary decision was then made on whether to discard this timeline or retain it to be annotated and thereby create a ground-truth timeline using it. This decision was based on a time-series visualization of the frequency of daily posts for that user and highlighting the location of the timeline to be either retained or discarded. The decision to discard a timeline was based on two criteria: whether the timeline (1) was primarily sparse over the full 15 days of the timelines, or to a lesser degree (2) whether it appeared that the CMoC was generated by noise. It was chosen to discard timelines that were (1) primarily sparse, to ensure that we allow sufficient amount of time to pass between posts such that moments of change can occur. Timelines that appeared to be (2) generated by noise, were discarded such that the ground-truth timelines were representative of timelines that would be generated by a change-point detection algorithm with well chosen hyper-parameters -- as the retained timelines were thus timelines that appeared to be generated by realistic change-points. Figure \ref{fig:discarded_timeline} presents a visualisation of a timeline that was discarded as described above, and figure \ref{fig:retained_timeline} describes a timeline that was included to be annotated as a ground-truth timeline.

This process of visually deciding whether a randomly sampled candidate timeline should be retained to be converted into a ground-truth timeline was repeated until 500 candidate timelines were retained. This process thus lasted until 1,220 randomly sampled timelines were observed and thus 720 timelines were discarded. 

From the annotated timelines, medoids are returned as the medoid timestamp of the annotated GTMoC after annotations were union aggregated across all annotators as described in \cite{tsakalidis2022MoC}.

\begin{figure}[!h]
    \centering
    \includegraphics[width=0.45\textwidth]{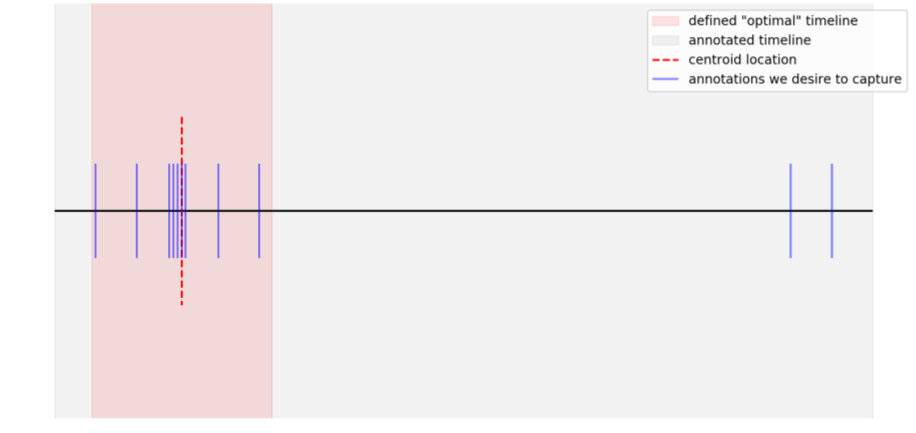}
    \caption{Identifying the position of the medoid, from the timestamps of posts annotated as GTMoCs.}
    \label{fig:Illustration for identifying the position of the centroid}
\end{figure}

\subsection{Annotation Guidelines}\label{appendix: annotation guidelines}

The annotation task proposed by \cite{tsakalidis2022MoC} was to assign annotators to identify changes in mood, by reading through the posts in chronological order included within the generated timeline of an individual -- and annotating the posts which contain a change in the user's mood compared to the recent past. 



\begin{figure}[!h]
    \centering
    \includegraphics[width=0.4\textwidth]{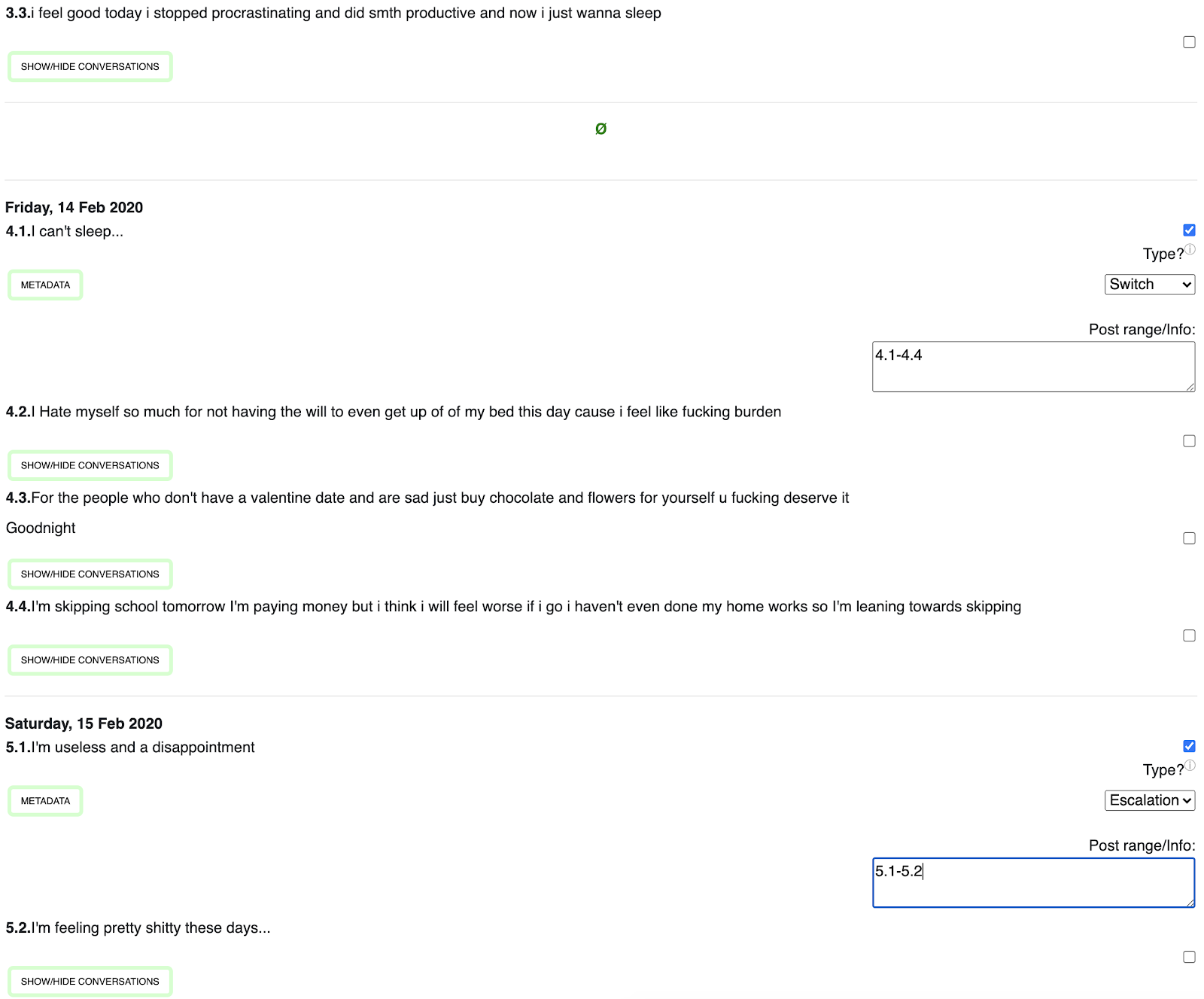}
    \caption{An example of the annotation interface, displaying a sequence of posts in a timeline shown to an annotator. For these sequence of posts, the annotator annotated a single post as a "switch" and another post as an "escalation". The user has a "switch" at 4.1, drastically changing from a positive mood to a negative mood -- where this changed mood persists until 4.4. The "escalation" begins and is at its peak (in this case becoming increasingly negative) at 5.1, and de-escalates up to the post at 5.2."}
    \label{fig:switch escalation}
\end{figure}

An example illustrating both a switch, and an escalation are displayed in figure \ref{fig:switch escalation}. Note, that the example shown in this figure will be paraphrased before the work is published -- to further preserve anonymity of this user.

\end{document}